\documentclass{article}

\usepackage{microtype}
\usepackage{graphicx}
\usepackage{booktabs} 
\usepackage{subcaption}
\usepackage{hyperref}

\usepackage[accepted]{icml2025}

\usepackage{amsmath}
\usepackage{amssymb}
\usepackage{mathtools}
\usepackage{amsthm}

\usepackage[capitalize,noabbrev]{cleveref}

\theoremstyle{plain}

\theoremstyle{definition}

\theoremstyle{remark}

\usepackage[textsize=tiny]{todonotes}

\begin{document}

\twocolumn[
\icmltitle{Reasoning-Finetuning Repurposes Latent Representations in Base Models}



\icmlsetsymbol{equal}{*}

\begin{icmlauthorlist}
\icmlauthor{Jake Ward}{equal,inst1}
\icmlauthor{Chuqiao Lin}{equal,inst2}
\icmlauthor{Constantin Venhoff}{inst3}
\icmlauthor{Neel Nanda}{}
\end{icmlauthorlist}

\icmlaffiliation{inst1}{Independent}
\icmlaffiliation{inst2}{Rudolf Peierls Centre for Theoretical Physics, Oxford}
\icmlaffiliation{inst3}{University of Oxford}

\icmlcorrespondingauthor{Jake Ward}{jakenicholasward@gmail.com}
\icmlcorrespondingauthor{Chuqiao Lin}{chuqiao.lin@physics.ox.ac.uk}

\icmlkeywords{Machine Learning, ICML}

\vskip 0.3in
]



\printAffiliationsAndNotice{\icmlEqualContribution}


\begin{abstract}
Backtracking, an emergent behavior elicited by reasoning fine-tuning, has been shown to be a key mechanism in reasoning models' enhanced capabilities. Prior work has succeeded in manipulating this behavior via steering vectors, but the underlying mechanism remains poorly understood. In this work, we show that the emergence of backtracking in DeepSeek-R1-Distill-Llama-8B is in part driven by a repurposed direction \emph{already present in base model activations}. Specifically, we identify a direction in base Llama-3.1-8B's residual stream which systematically induces backtracking when used to steer the distilled reasoning model, and find that the effects of steering with this direction cannot be trivially explained by token-level attributes. We further find that this direction \emph{does not} induce backtracking in the base model, suggesting that the reasoning finetuning process repurposes pre-existing representations to form new behavioral circuits. Additionally, we hypothesize that this direction is one of several which may work together to mediate backtracking. Our findings offer a compelling picture that reasoning-finetuned models repurpose pre-existing base model representations, rather than learn new capabilities from scratch.
\end{abstract}

\section{Introduction}
\label{intro}

Recent advancements in large language model (LLM) post-training has led to a new class of \emph{reasoning models}, which can leverage test-time compute to achieve enhanced performance on reasoning-intensive benchmarks \cite{wei2022chain, guo2025deepseek, yeo2025demystifying, yang2025step, ye2025emergence}. These models tend to exhibit an emergent behavior often referred to as \emph{backtracking}, where after progressing down a reasoning path or coming up with a candidate answer, a model will explore alternative strategies \cite{venhoff2025understanding}. Empirically, the presence of backtracking accounts for a substantial fraction of the accuracy gap between base models and their reasoning-fine-tuned counterparts \cite{muennighoff2025scaling}.

Prior work has shown that this behavior can be reliably induced using \emph{steering vectors} derived from activation differences at sentences classified as backracking \cite{venhoff2025understanding}. While Venhoff et al. have shown that steering vectors can be used to control backtracking behavior, the fundamental mechanism underlying this behavior remains poorly understood.

\begin{figure*}
    \centering
    \includegraphics[width=1.0\linewidth]{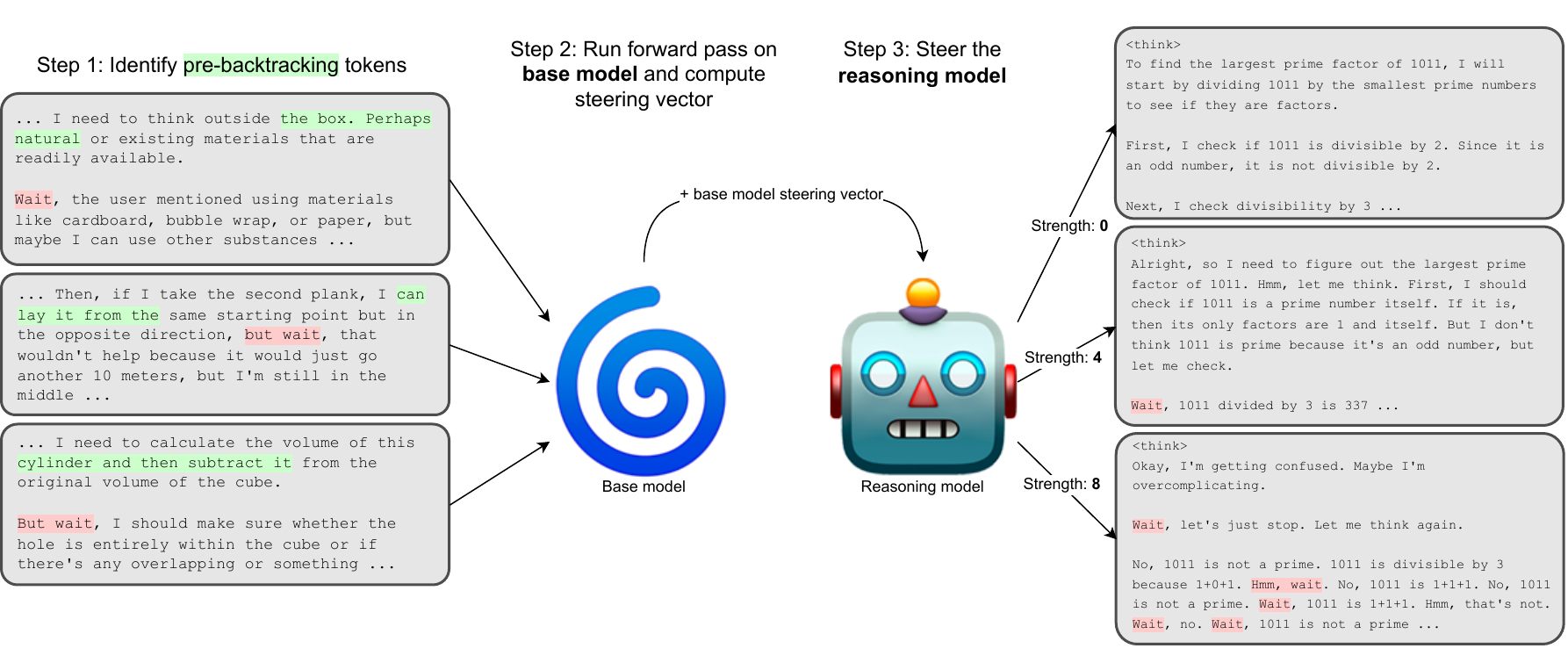}
    \caption{Steering vectors derived from base model activations induce backtracking when used to steer the reasoning-finetuned model. Green highlights represent tokens from which our backtracking steering vectors are computed, red highlights indicate the start of backtracking.}
    \label{fig:nice_illustration}
\end{figure*}

In this work, we perform a deeper investigation into backtracking steering vectors and investigate how and where they emerge in model activations. Concretely, we find that a backtracking steering vector can be computed using (1) activations at an offset token position \textit{preceding} the backtracking event, allowing capture of upstream, causally relevant concepts, and (2) activations sampled only from the \textit{base model}, suggesting that the underlying mechanism of backtracking partially emerges from a concept already represented in base model activations (Fig. \ref{fig:nice_illustration}). Crucially, while this representation is shared by both base and reasoning models, the representation only induces backtracking in the reasoning model, suggesting that it has been repurposed as an input to the backtracking mechanism during reasoning finetuning.

In Sec. \ref{sec:steering_vector_analysis}, we show that backtracking steering vectors derived from \textit{base model} activations reliably invoke backtracking when used to steer the \textit{reasoning model} (Fig. \ref{fig:nice_illustration}). In Sec. \ref{sec:non_trivial_concepts}, we use logit lens and probing to show that the backtracking-inducing direction is non-trivial: it does not directly boost the logits of backtracking-related keywords (e.g. ``Wait''). We additionally find this direction is densely present in model activations across contexts, suggesting that it is not the sole factor mediating backtracking behavior.

Taken together, our findings sharpen our understanding of how reasoning capabilities emerge in finetuned models. Rather than learning reasoning behavior entirely from scratch, these models respond to and repurpose signals already present in the base models. This suggests that base models may possess latent reasoning capabilities which are unexpressed until they are extracted by the finetuning process. Broadly, we hope this work inspires focused, fine-grained investigations into various aspects of reasoning processes through the lens of interpretability.

\section{Preliminaries}

\subsection{Deriving steering vectors}
We derive steering vectors using the \textbf{Difference-of-Means} (DoM) method \cite{rimsky-etal-2024-steering, venhoff2025understanding}. Full derivation and details are provided in Appendix \ref{app:steering_vectors} for completeness.





\subsection{Detection of backtracking}

Following \cite{venhoff2025understanding}, we use a GPT-4o judge to identify backtracking events in 300 reasoning model output traces. Separately, we operationalize backtracking as the fraction of output tokens matching a predefined keyword set (e.g. ``Wait'' or ``But''), which has been shown to coincide with backtracking \cite{galichin2025have}. We define this proxy metric as:
\begin{equation}
\label{eq:backtrack_score}
    b := \frac{1}{N}\sum_{\mathrm{word} \in \substack{\mathrm{reasoning} \\ \mathrm{trace}}} \mathbb{I}[{\mathrm{word} \in \mathcal{B}}]
\end{equation}
where $N$ counts the number of words in the reasoning trace, and $\mathcal{B}= \{\texttt{wait}, \texttt{hmm}\}$ is the set of backtracking keywords. Further justifications of the metric are provided in Appendix \ref{app:consistency}, where we explore the consistency between LLM (GPT-4o) judges, keyword detection, and human judges in identifying whether a reasoning output component is backtracking or not. We demonstrate that our keyword metric is an acceptable indicator of backtracking events for our purposes.


\section{Steering vector analysis}
\label{sec:steering_vector_analysis}



Extending previous work on steering vectors and reasoning models \cite{venhoff2025understanding}, we study steering vectors with two key properties: (1) \textbf{Negative token offset}: In addition to deriving backtracking steering vectors from token positions where backtracking occurs, we use token positions with a negative offset from the actual backtracking event. The fact that these vectors are derived from positions before backtracking occurs suggests that these directions are causally relevant to the model's decision to backtrack. (2) \textbf{Using base model activations}: Beyond sampling activations from the finetuned reasoning model only, we derive steering vectors separately on residual stream activations from both the \textit{base} and \textit{reasoning} models on the same reasoning traces. We refer to these as ``\textit{base-derived}'' and ``\textit{reasoning-derived}'' steering vectors where appropriate.

\subsection{Deriving steering vectors with a negative offset}

\begin{figure}[h]
    \centering
    \includegraphics[width=1.0\linewidth]{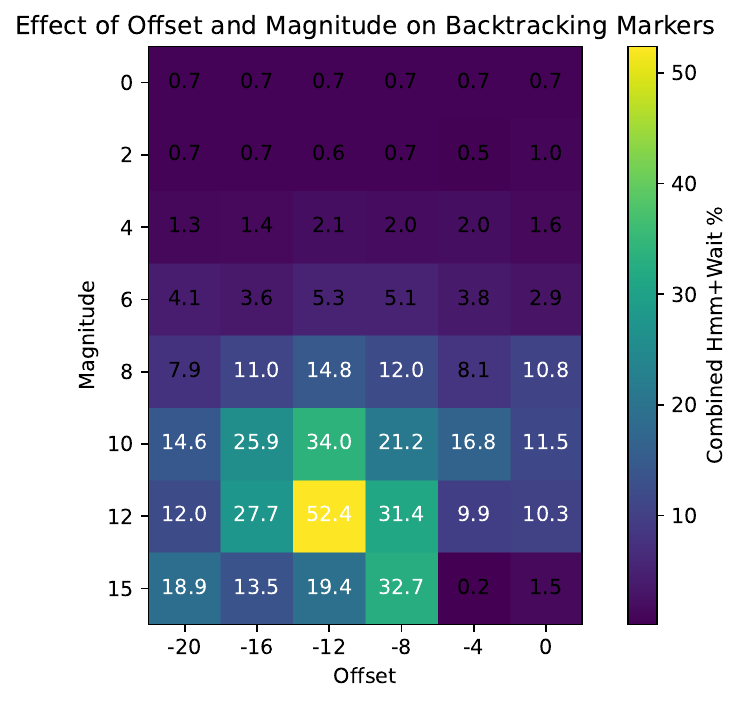}
    \caption{The effect of steering as a function of token window offset and steering vector magnitude. Steering vectors are derived from the layer 10 residual stream of the reasoning model. The colorbar reflects the metric value from Eq. \eqref{eq:backtrack_score} averaged over multiple generated steered reasoning traces.}
    \label{fig:optim_steering_offset}
\end{figure}

Fig. \ref{fig:optim_steering_offset} measures the effectiveness of backtracking steering vectors across different steering offsets and magnitudes. We find that the optimal offset for $10^\mathrm{th}$-layer residual stream of DeepSeek-R1-Distill-Llama-8B is $\sim -13$ to $-8$. This window usually covers the beginning of the sentence prior to backtracking. We also show that these optimally derived steering vectors outperform the no-offset baseline. We fix this offset for the remainder of our analysis.

\subsection{Using base model activations}

\begin{figure}[h]
    \centering
    \includegraphics[width=1\linewidth]{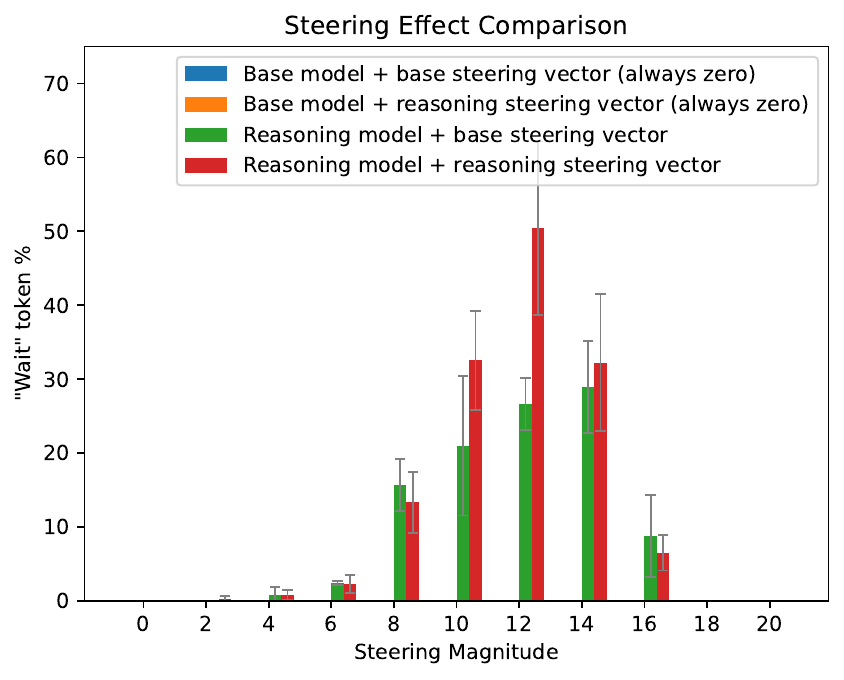}
    \caption{Proportion of backtracking-related tokens generated by both base and reasoning models when steered with base-derived or reasoning-derived steering vectors. Gray lines represent error bars of one standard deviation. Note that the base model never exhibits backtracking behavior, even when steered with the reasoning model-derived backtracking-inducing vector.}
    \label{fig:base_and_ft_trained_steering_vec}
\end{figure}

We compute steering vectors separately from base and reasoning model residual stream activations at layer 10, and examine the effect these vectors have when added to each models' activations during generation. In Fig. \ref{fig:base_and_ft_trained_steering_vec} we report the striking result that \textit{base}-derived steering vectors reliably induce backtracking when used to steer the \emph{reasoning} model, and have comparable performance to their reasoning-derived counterparts. Additionally, we find that neither base-derived nor reasoning-derived steering vectors invoke backtracking behavior in the base model.

We additionally examine backtracking steering vectors computed at each of the 32 layers in the studied models, and find both base-derived and reasoning-derived steering vectors are most effective around layer 10. Readers are referred to Appendix \ref{app:steering_vectors_all_layers} and Fig. \ref{fig:steering_all_layers} for more detailed presentations. 

We find that base- and reasoning-derived steering vectors have high cosine similarity of $\sim0.74$, suggesting that they are capturing the same representation. While the existence of effective, mostly parallel steering vectors strongly indicates a \textit{shared representation} between base and reasoning models, we find that only the reasoning model uses this representation to initiate backtracking (Fig. \ref{fig:base_and_ft_trained_steering_vec}). In light of this, we conjecture that the extracted backtracking steering vectors may actually represent some more abstract concept, and that base and reasoning models use this concept differently for downstream generation.

\subsection{Validation against baselines}

To ensure that the increased backtracking behavior we observe is not merely a consequence of \textit{any} perturbation in activations, we compare the effect of our steering vector to that of various baselines: (1) \textbf{overall mean} - adding the mean activation to activations; (2) \textbf{noise} - adding random Gaussian noise; (3) \textbf{self-amplification} - increasing activation magnitudes by adding each activation to itself scaled by a coefficient; (4) \textbf{deduction} - steering vectors from tokens labeled as deductive reasoning steps; and (5) \textbf{initializing} - steering vectors from problem setup tokens.

\begin{figure}[h]
    \centering
    \includegraphics[width=1.0\linewidth]{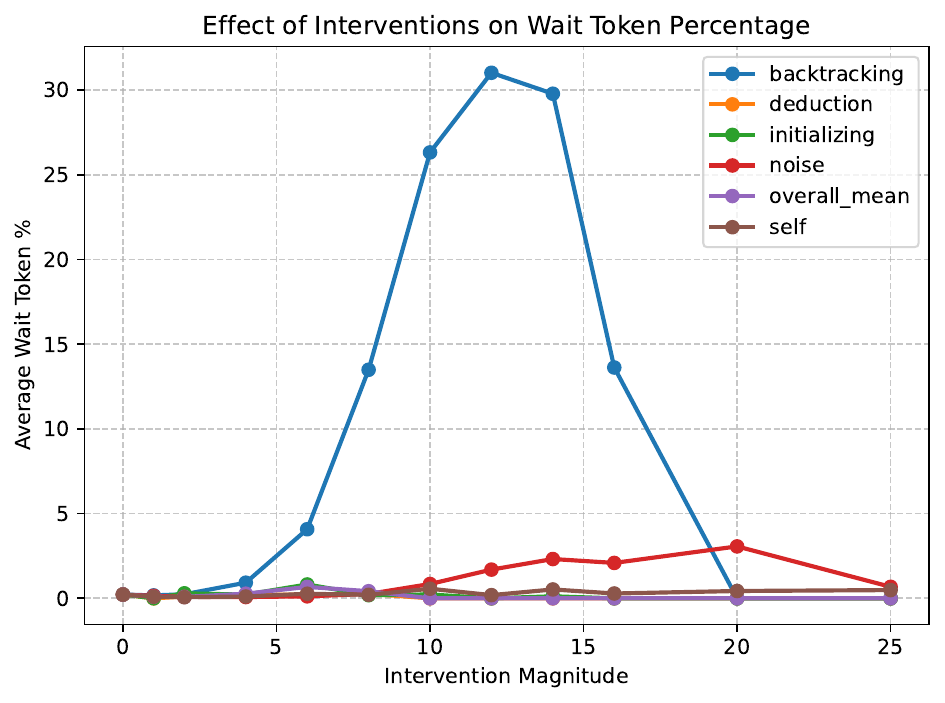}
    \caption{Comparison of various baselines used to steer the reasoning model, measured by the ``Wait'' metric. The base model-derived, negative-offset backtracking steering vector (ours) clearly has a significant effect.}
    \label{fig:baseline_comparison_main}
\end{figure}

As shown in Fig. \ref{fig:baseline_comparison_main}, the backtracking steering vector significantly outperforms all tested baselines, validating that we have identified a meaningful direction rather than an artifact of arbitrary perturbations.

Notably, adding Gaussian noise to activations has a nontrivial effect on the fraction of output words which are ``Wait''. We observe anecdotally that this type of intervention results in coherent outputs with increased propensity for backtracking. We leave investigation of this phenomenon as a direction for future work.

\section{Steering vector directions represent nontrivial concepts}
\label{sec:non_trivial_concepts}
In this section, we investigate what our identified backtracking-inducing vectors encode. Naively, one might expect these vectors to simply boost the output logit probability for trigger tokens like ``wait'' \cite{muennighoff2025scaling}. We use logit lens to show that token-level attributes cannot explain backtracking behavior in our steering vectors. Given the effectiveness of our base-derived vector, optimists might anticipate (in view of \textit{refusal} directions \cite{arditi2024refusal}) that backtracking is mediated by this single direction. However, probing experiments reveal that the backtracking-inducing direction we find is not alone sufficient to fully explain backtracking behavior.

\subsection{Logit lens}
One possible trivial explanation for the backtracking-inducing behavior of our steering vectors is that they merely boost the probability of backtracking-related tokens via a direct projection onto the unembedding directions for these tokens. We refute this explanation by showing that the base-derived vector \emph{does not} have significant positive projections onto the relevant unembed directions. We compute a ``backtracking score'' $s$ by computing the projection of our vectors onto the unembed matrix, with directions for irrelevant tokens (non-backtracking-related) masked out:

\begin{equation}
    s(\mathbf{v}) := (W_U \mathbf{v}) \cdot \frac{\mathbf{a}}{\|\mathbf{a}\|_1}, a_i = \left\{\begin{array}{cc}
       1  &  \text{if $\mathrm{Decode}(i)\in \mathcal{B}$} \\
       0  &  \text{otherwise}
    \end{array}\right.
\end{equation}

where $\mathbf{v}$ is the steering vector, $W_U$ is the unembedding matrix of either base or fine-tuned model and $\mathbf{a}$ is a mask which selects for backtracking keywords. In this section, we use $\mathcal{B} = \{\texttt{wait},\texttt{but}\}$ \footnote{This covers all tokens in the vocabulary which contain ``wait'' or ``but'', case insensitive. Examples: \texttt{\_Wait}, \texttt{\_wait}, \texttt{Wait}, etc.}. 

\begin{figure}[h]
    \centering
    \includegraphics[width=\linewidth]{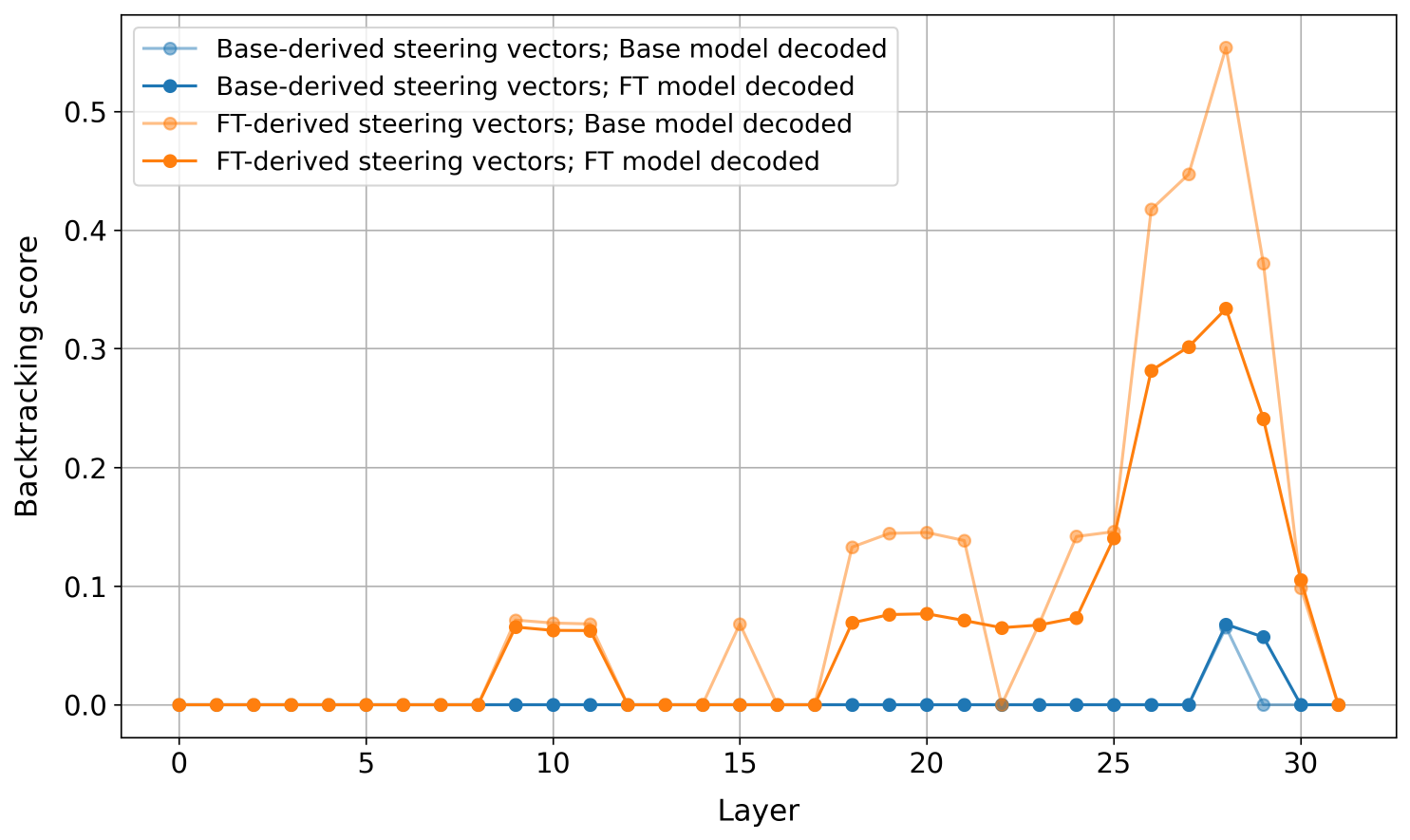}
    \caption{Backtracking scores of steering vectors trained on base model (blue) or reasoning-finetuned model (orange) activations at different layers, when projected onto base (light) or reasoning (dark) model unembedding matrices.}
    \label{fig:logit_lens}
\end{figure}

Fig. \ref{fig:logit_lens} reports the backtracking scores of base- and reasoning-derived steering vectors computed from hidden activations at different layers. Combined with findings in the previous section, we observe that (1) the base-derived steering vectors do not decode to backtracking keywords, yet are successful in eliciting backtracking in the fine-tuned model; (2) although later-layer steering vectors can be attributed to token-level logit boosts, they are less effective when used for steering. From this, we claim that our layer-10 base-derived steering vector is capturing a more abstract concept, causally relevant for backtracking.

\subsection{Probing}

To better understand our identified backtracking-inducing direction, we conduct case studies in which we use this direction for probing. We examine the magnitude of the projection of these vectors onto centered model activations and attempt to identify an interpretable semantic meaning.

\begin{figure}[h]
    \centering
    \includegraphics[width=1.0\linewidth]{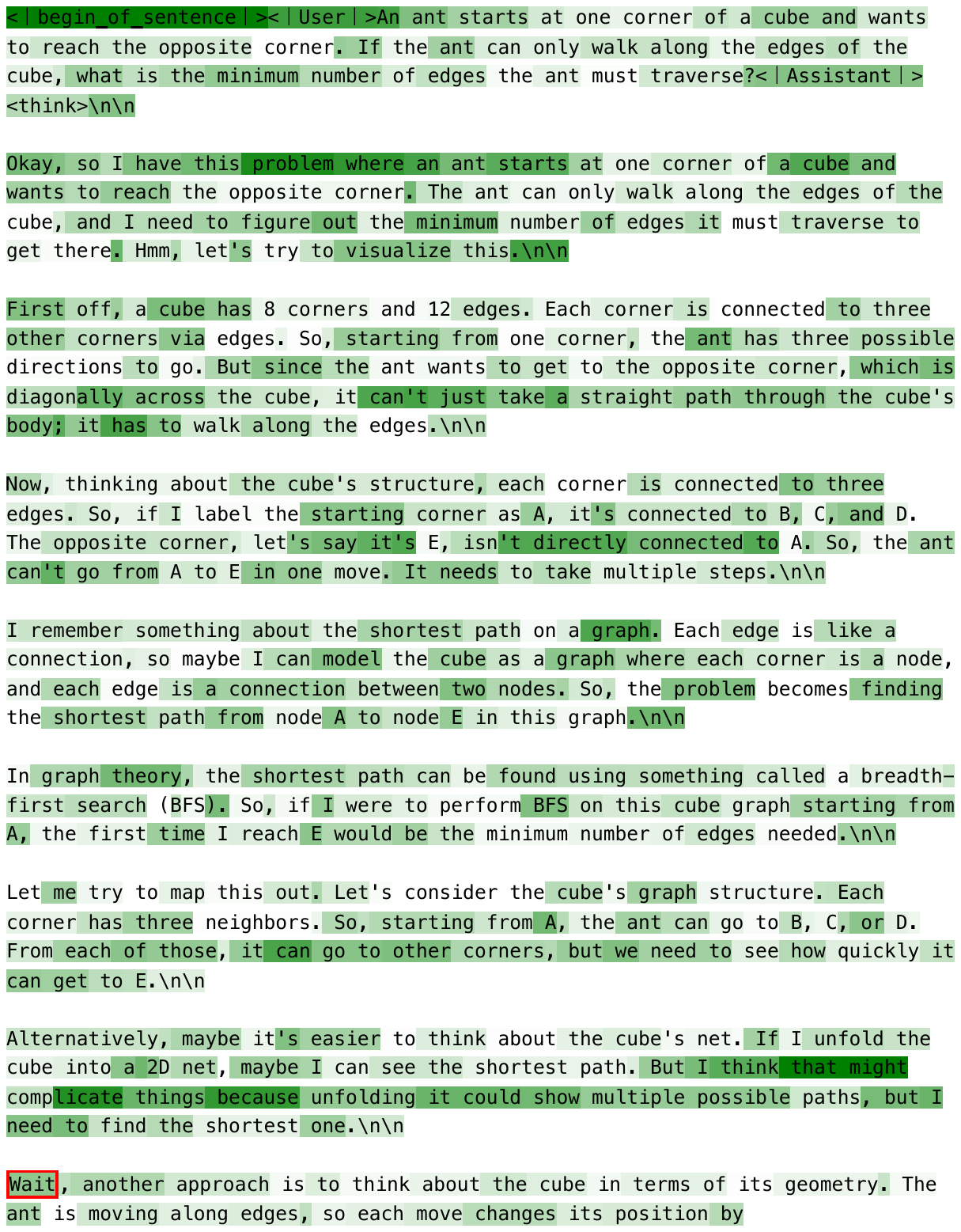}
    \caption{A sample output generated by the reasoning model without steering. Tokens are highlighted in green when the projection of the base model-derived backtracking steering vector onto layer 10 activations is positive, with darker green representing higher magnitudes. ``Wait'' tokens outlined in red for clarity.}
    \label{fig:probing}
\end{figure}

We find that, unexpectedly, the base-derived steering direction is densely present in model activations (Fig. \ref{fig:probing}), and that it does not cleanly correlate with backtracking when used as a probe. These results lead us to hypothesize that our identified direction may be one of \textit{several} heuristics the reasoning model uses to trigger backtracking. Our observation that our identified direction is effective at moderate (but not small) steering strengths indicates that the actual backtracking mechanism may involve a linear combination of such heuristics. We leave further investigation of this hypothesis to future work.

\section{Conclusion and Outlook}

In this work, we have provided evidence that the emergent backtracking behavior expressed by the DeepSeek-R1-Distill-Llama-8B reasoning model arises, at least in part, through the repurposing of pre-existing representations in the base model from which it was finetuned. More specifically, we have identified a direction present in Llama-3.1-8B residual stream activations which, when added to reasoning model activations, systematically induces backtracking. The fact that this representation exists in the base model without inducing backtracking there offers a key insight: that the reasoning finetuning process may elicit backtracking behavior by repurposing latent representations already present in the base model, rather than learning the entire mechanism from scratch.

Our analysis has several limitations. First, we examined a single reasoning model; further investigation to validate the robustness of our findings across different model families and scales is required in order to claim our findings are general. Second, our identified steering direction is densely present in model activations and we find instances both where backtracking occurs while the direction is not present, and instances where the direction is present but backtracking does not occur. This indicates that we have only identified one component of the backtracking mechanism, and the remainder of the mechanism remains unknown. Our results should be interpreted primarily as an existence proof for latent reasoning-related representations in base models, rather than a comprehensive explanation of reasoning behavior.

Our work highlights the value of interpretability tools in uncovering nuanced insights regarding the mechanisms underlying emergent capabilities in LLMs. We hope that these insights will ultimately lead to more transparent and controllable artificial intelligence.

\section*{Acknowledgements}

We acknowledge the support of the MATS 8.0 program during which foundational experiments for this work were completed. C.L. thanks Arthur Conmy and Shivam Raval for useful comments on related research topics, and Shivaji Sondhi and The Leverhulme Trust for kind support provoiding GPU compute, and introduction to the field of interpretability more generally.

\bibliography{ref}

\begin{thebibliography}{10}
\providecommand{\natexlab}[1]{#1}
\providecommand{\url}[1]{\texttt{#1}}
\expandafter\ifx\csname urlstyle\endcsname\relax
  \providecommand{\doi}[1]{doi: #1}\else
  \providecommand{\doi}{doi: \begingroup \urlstyle{rm}\Url}\fi

\bibitem[Arditi et~al.(2024)Arditi, Obeso, Syed, Paleka, Panickssery, Gurnee, and Nanda]{arditi2024refusal}
Arditi, A., Obeso, O., Syed, A., Paleka, D., Panickssery, N., Gurnee, W., and Nanda, N.
\newblock Refusal in language models is mediated by a single direction.
\newblock \emph{arXiv preprint arXiv:2406.11717}, 2024.

\bibitem[Galichin et~al.(2025)Galichin, Dontsov, Druzhinina, Razzhigaev, Rogov, Tutubalina, and Oseledets]{galichin2025have}
Galichin, A., Dontsov, A., Druzhinina, P., Razzhigaev, A., Rogov, O.~Y., Tutubalina, E., and Oseledets, I.
\newblock I have covered all the bases here: Interpreting reasoning features in large language models via sparse autoencoders.
\newblock \emph{arXiv preprint arXiv:2503.18878}, 2025.

\bibitem[Guo et~al.(2025)Guo, Yang, Zhang, Song, Zhang, Xu, Zhu, Ma, Wang, Bi, et~al.]{guo2025deepseek}
Guo, D., Yang, D., Zhang, H., Song, J., Zhang, R., Xu, R., Zhu, Q., Ma, S., Wang, P., Bi, X., et~al.
\newblock Deepseek-r1: Incentivizing reasoning capability in llms via reinforcement learning.
\newblock \emph{arXiv preprint arXiv:2501.12948}, 2025.

\bibitem[Niklas~Muennighoff(2025)]{muennighoff2025scaling}
Niklas~Muennighoff, Zitong~Yang, W. S. X. L. L. L. F.-F. H. H. L. Z. P. L. E. C. T.~H.
\newblock s1: Simple test-time scaling.
\newblock \emph{arXiv preprint arXiv:2501.19393}, 2025.

\bibitem[Rimsky et~al.(2024)Rimsky, Gabrieli, Schulz, Tong, Hubinger, and Turner]{rimsky-etal-2024-steering}
Rimsky, N., Gabrieli, N., Schulz, J., Tong, M., Hubinger, E., and Turner, A.
\newblock Steering llama 2 via contrastive activation addition.
\newblock In Ku, L.-W., Martins, A., and Srikumar, V. (eds.), \emph{Proceedings of the 62nd Annual Meeting of the Association for Computational Linguistics (Volume 1: Long Papers)}, pp.\  15504--15522, Bangkok, Thailand, August 2024. Association for Computational Linguistics.
\newblock \doi{10.18653/v1/2024.acl-long.828}.
\newblock URL \url{https://aclanthology.org/2024.acl-long.828/}.

\bibitem[Venhoff et~al.(2025)Venhoff, Arcuschin, Torr, Conmy, and Nanda]{venhoff2025understanding}
Venhoff, C., Arcuschin, I., Torr, P., Conmy, A., and Nanda, N.
\newblock Understanding reasoning in thinking language models via steering vectors.
\newblock In \emph{Workshop on Reasoning and Planning for Large Language Models}, 2025.
\newblock URL \url{https://openreview.net/forum?id=OwhVWNOBcz}.

\bibitem[Wei et~al.(2022)Wei, Wang, Schuurmans, Bosma, Xia, Chi, Le, Zhou, et~al.]{wei2022chain}
Wei, J., Wang, X., Schuurmans, D., Bosma, M., Xia, F., Chi, E., Le, Q.~V., Zhou, D., et~al.
\newblock Chain-of-thought prompting elicits reasoning in large language models.
\newblock \emph{Advances in neural information processing systems}, 35:\penalty0 24824--24837, 2022.

\bibitem[Yang et~al.(2025)Yang, Zhu, Wei, Zhang, Shao, Zhou, Guo, and Li]{yang2025step}
Yang, X.-W., Zhu, X.-Y., Wei, W.-D., Zhang, D.-C., Shao, J.-J., Zhou, Z., Guo, L.-Z., and Li, Y.-F.
\newblock Step back to leap forward: Self-backtracking for boosting reasoning of language models.
\newblock \emph{arXiv preprint arXiv:2502.04404}, 2025.
\newblock URL \url{https://arxiv.org/abs/2502.04404}.

\bibitem[Ye et~al.(2025)Ye, Pham, Zhang, Gopi, Peng, Li, Kulkarni, and Inan]{ye2025emergence}
Ye, G., Pham, K.~D., Zhang, X., Gopi, S., Peng, B., Li, B., Kulkarni, J., and Inan, H.~A.
\newblock On the emergence of thinking in llms i: Searching for the right intuition.
\newblock \emph{arXiv preprint arXiv:2502.06773}, 2025.
\newblock URL \url{https://arxiv.org/abs/2502.06773}.

\bibitem[Yeo et~al.(2025)Yeo, Lin, Han, Lin, Wang, Wang, Zhang, and Zhuo]{yeo2025demystifying}
Yeo, E.-H., Lin, Z., Han, T., Lin, M., Wang, J., Wang, J., Zhang, J., and Zhuo, D.
\newblock Demystifying long chain-of-thought reasoning in llms.
\newblock \emph{arXiv preprint arXiv:2502.03373}, 2025.
\newblock URL \url{https://arxiv.org/abs/2502.03373}.

\end{thebibliography}
\bibliographystyle{icml2025}

\newpage
\appendix
\renewcommand{\thefigure}{\Alph{section}.\arabic{figure}}
\renewcommand{\thetable}{\Alph{section}.\arabic{table}}
\numberwithin{figure}{section}
\numberwithin{table}{section}
\onecolumn

\section{Deriving steering vectors}
\label{app:steering_vectors}
In this work, we derive steering vectors using the \textbf{Difference-of-Means} (DoM) method \cite{rimsky-etal-2024-steering, venhoff2025understanding}. We use DeepSeek-R1-Distill-Llama-8B to generate a set of tokenized reasoning traces $\mathcal{R}$. We then extract a dataset $\mathcal{D}= \{(p_i,S_i)| i \in [|\mathcal{R}|]\}$, where $p_i$ is the tokenized prompt of the $i^\mathrm{th}$ reasoning trace in $\mathcal{R}$, and $S_i$ is a subset of sequence positions where we would like to train the steering vector (e.g. fixed offset preceding sentence terminations). We further extract $\mathcal{D}_+ \subset \mathcal{D}$ where the backtracking-related token is present, as annotated by GPT-4o. The steering vector $\mathbf{v}$ is computed as
\begin{equation}
\begin{aligned}
\label{eq:DiM}
& \mathrm{MeanAct}(\mathcal{D}) := \frac{1}{|\mathcal{D}|} \sum_{i=1}^{|\mathcal{D}|} \frac{1}{|S_i|} \sum_{s\in S_i} A(p_i)[s] \
& \mathbf{v} = \mathrm{MeanAct}(\mathcal{D}_+) - \mathrm{MeanAct}(\mathcal{D})
\end{aligned}
\end{equation}
where $A(p_i)$ is the target hidden state (at some target layer) when forward-passing the tokenized prompt $p_i$, and $[s]$ accesses the subset sequence positions.

Following the methodology of \cite{venhoff2025understanding}, we generate a dataset for computing steering vectors:

\begin{enumerate}
    \item We generate 300 prompts within 10 different categories using Claude Sonnet 3.7.
    \item We use DeepSeek-R1-Distill-Llama-8B to generate reasoning traces for these prompts.
    \item We use GPT-4o to classify sentences in the generated reasoning traces as ``backtracking'' or otherwise.
\end{enumerate}

\section{More on training steering vectors}
\label{app:more_steering_vectors}
\subsection{Base-derived and reasoning-derived steering vectors from all hidden layers}
\label{app:steering_vectors_all_layers}

\begin{figure}[h]
    \centering
    \begin{subfigure}[t]{0.48\linewidth}
        \centering
        \includegraphics[width=\linewidth]{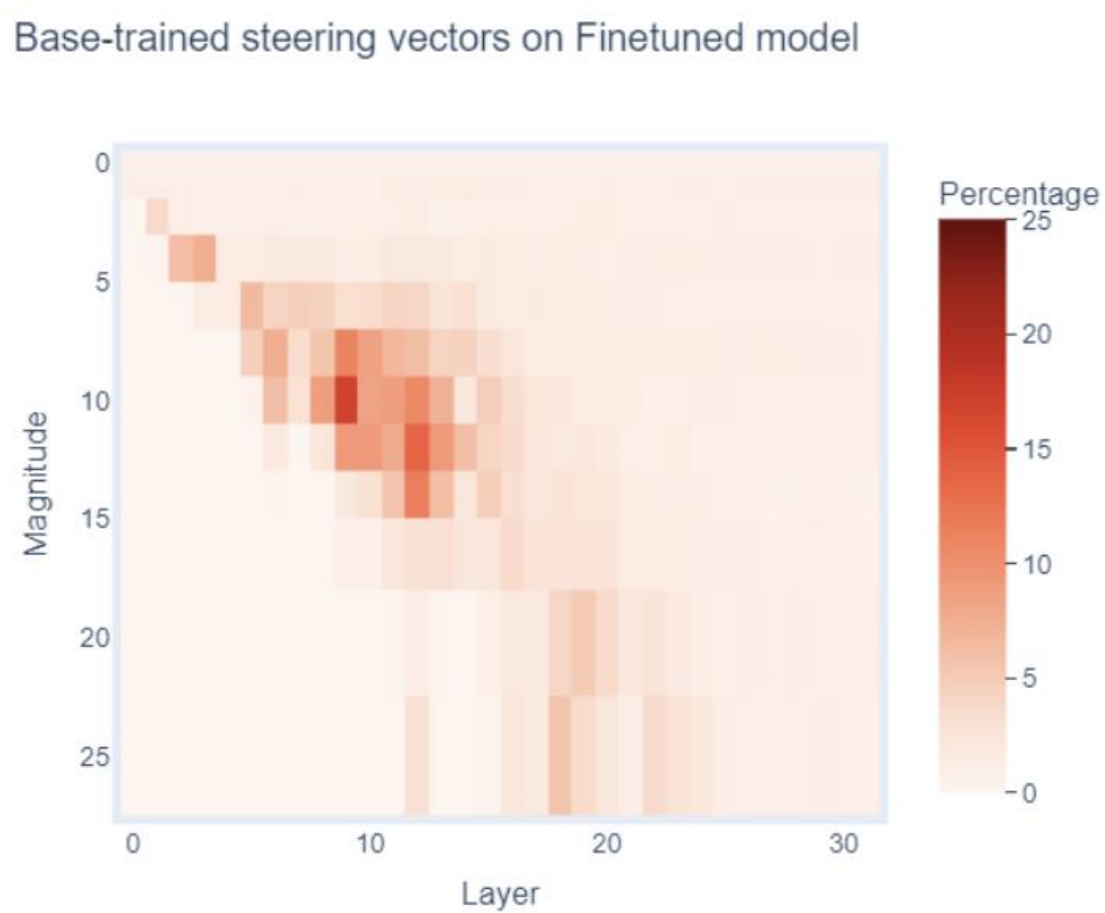}
        \caption{Downstream backtracking behavior response to base-derived steering vectors, as measured by proportion of output tokens in the steered reasoning trace which are backtracking-related.}
        \label{fig:subfig-a}
    \end{subfigure}
    \hfill
    \begin{subfigure}[t]{0.48\linewidth}
        \centering
        \includegraphics[width=\linewidth]{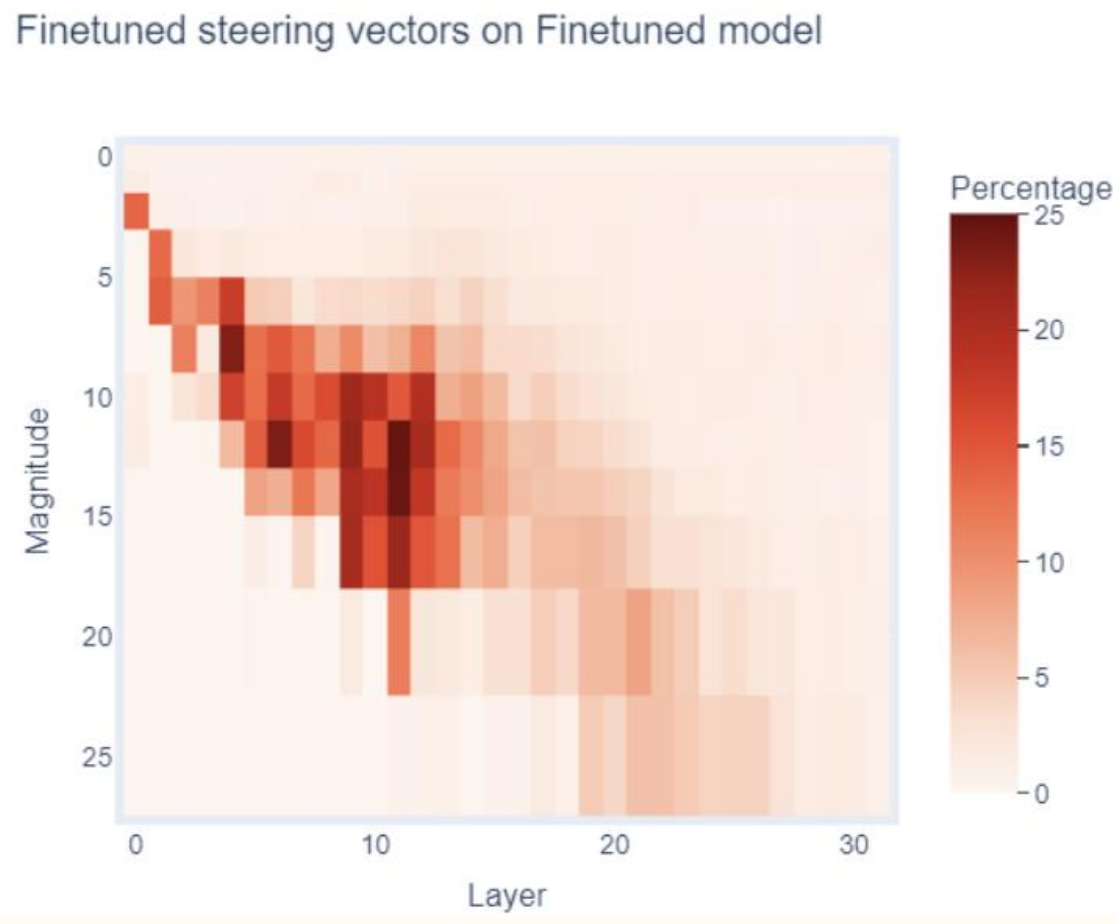}
        \caption{Downstream backtracking behavior response to reasoning-derived steering vectors, as measured by proportion of output tokens in the steered reasoning trace which are backtracking-related.}
        \label{fig:subfig-b}
    \end{subfigure}
    \caption{Comparison of steering effectiveness when training on base vs. fine-tuned model activations.}
    \label{fig:steering_all_layers}
\end{figure}

\section{Consistency between human, LLM and keyword judges for backtracking detection}
\label{app:consistency}

Throughout our analyses, we employ a keyword counting metric to measure the presence of backtracking in various steering regimes. We refer to this metric here as the ``keyword judge''. This metric classifies a reasoning trace component as ``backtracking'' by the presence of keyword tokens. We believe this is a reasonable proxy for detecting backtracking behavior.

\paragraph{LLM judge}
When generating a labeled dataset for deriving steering vectors, we prompt GPT-4o to annotate reasoning traces based on a prescribed sentence taxonomy c.f. \cite{venhoff2025understanding}.

\paragraph{Keyword judge}
As explained above, to judge whether a reasoning trace component is backtracking or not, we look for the pattern \texttt{wait} in the decoded text. 

In Table \ref{tab:consistency_keyword_llm}, we analyze the consistency between LLM and keyword judges when classifying reasoning sentences generated by the fine-tuned model with and without backtracking steering, and at various steering strengths. While backtracking occurs in fewer than two percent of sentences on average, we see F1 scores above 60\% at intermediate steering strengths, showing that we have reasonable agreement between detections made by either LLM or keyword judge.

On the other hand, the numbers still signal a nontrivial gap between the classification decisions made by LLM judge and the keyword judge - To further resolve such discrepancies, we introduce ourselves as human judges for backtracking ground truth. Extensive case study further unravels inadequacy of LLM and keywords as judges. Quantitatively, we find that $\sim 83\%$ of sentences identified with the keyword metric are true examples of backtracking when we use our own judgement as the ground truth. Qualitatively, we find that some of the discrepancy between keyword and LLM judges stems from the LLM classifiying backtracking sentences as something other than backtracking, like ``uncertainty-estimation''.

In summary, we believe the LLM judge and keyword metrics are both reasonable proxies for ``true'' backtracking for our purposes. However, applications where high precision and accuracy is critical will require more sophisticated metrics.

\begin{table}[t]
\vskip 0.15in
\begin{center}
\begin{small}
\begin{sc}
\begin{tabular}{lccccr}
\toprule
Steering Strength & Precision & Recall & F1-score \\
\midrule
0    & 63.36\% & 64.91\% & 64.12\% \\
4    & 62.50\% & 66.67\% & 64.52\% \\
8    & 48.11\% & 68.00\% & 56.35\% \\
12   & 47.66\% & 62.93\% & 54.24\% \\
\bottomrule
\end{tabular}
\end{sc}
\end{small}
\end{center}
\caption{Metrics of fit for keyword judge against LLM judge (treated as ground truth). The dataset we used consists of 300 questions featuring basic logic, geometry and probability.}
\label{tab:consistency_keyword_llm}
\vskip -0.1in
\end{table}





\end{document}